\documentclass{article}
\usepackage{ijcai22}

\usepackage{times}
\usepackage{amsfonts}
\usepackage{arydshln}
\usepackage{soul}
\usepackage{url}
\usepackage{color}
\usepackage[utf8]{inputenc}
\usepackage[small]{caption}
\usepackage{graphicx}
\usepackage{amsmath,amssymb}
\usepackage{amsthm}
\usepackage{booktabs}
\usepackage{algorithm}
\usepackage{algorithmic}
\usepackage{multirow}
\usepackage{bbding}
\usepackage{amssymb}
\usepackage{pifont}
\newcommand{\cmark}{\ding{51}}
\newcommand{\xmark}{\ding{55}} 

\title{Unsupervised Misaligned Infrared and Visible Image Fusion via Cross-Modality Image Generation and Registration}

\author{
    Di Wang$^{1}$\and
    Jinyuan Liu$^{1}$\and
    Xin Fan$^{2}$\And
    Risheng Liu$^{2}$\thanks{Corresponding author}
    \affiliations
    $^1$School of Software Technology, Dalian University of Technology, China\\
    $^2$International School of Information Science \& Engineering,
Dalian University of Technology, China\\
    \emails
    diwang1211@mail.dlut.edu.cn, atlantis918@hotmail.com,
    \{xin.fan, rsliu\}@dlut.edu.cn
}

\begin{document}

\maketitle

\begin{abstract}
%
Recent learning-based image fusion methods have marked numerous progress in pre-registered multi-modality data, but suffered serious ghosts dealing with misaligned multi-modality data, due to the spatial deformation and the difficulty narrowing cross-modality discrepancy.
To overcome the obstacles, in this paper, we present a robust cross-modality generation-registration paradigm for unsupervised misaligned infrared and visible image fusion (IVIF).
Specifically, we propose a Cross-modality Perceptual Style Transfer Network (CPSTN) to generate a pseudo infrared image taking a visible image as input.
Benefiting from the favorable geometry preservation ability of the CPSTN, the generated pseudo infrared image embraces a sharp structure, which is more conducive to transforming cross-modality image alignment into mono-modality registration coupled with the structure-sensitive of the infrared image. 
In this case, we introduce a Multi-level Refinement Registration Network (MRRN) to predict the displacement vector field between distorted and pseudo infrared images and reconstruct registered infrared image under the mono-modality setting.
Moreover, to better fuse the registered infrared images and visible images, we present a feature Interaction Fusion Module (IFM) to adaptively select more meaningful features for fusion in the Dual-path Interaction Fusion Network (DIFN).
Extensive experimental results suggest that the proposed method performs superior capability on misaligned cross-modality image fusion. 
%
\end{abstract}

\section{Introduction}
Image fusion is to extract the complementary information from different modality images and aggregate them to generate richer and more meaningful feature representations. Typically, Infrared and Visible Image Fusion (IVIF) enjoys the merit and presents conduciveness to practical applications such as autonomous driving and video surveillance.

Most existing IVIF methods~\cite{liu2020bilevel,liu2021searching,liu2022attention} are designed for the hand-crafted pre-registered images, which are very labor-intensive and time-sensitive, despite numerous progress.
%
And, they are sensitive to the discrepancy of intensity of the misaligned infrared and visible images, resulting in serious ghosting artifacts on fused images, once slight offset and deformation exist.
The intrinsic reason is that the large cross-modality variation between infrared and visible images makes it impractical to straightly bridge the domain gap between them in a shared feature space. Coupled with the absence of cross-modality similarity constraints, few attempts have ever been made to fuse the misaligned infrared and visible images. The primary obstacle is cross-modality image alignment.

More often than not, existing widely-used image alignment methods~\cite{flownet,voxelmorph} perform pixel-wise and feature-level alignment by explicitly estimating the deformation field between the distorted image and its reference. However, they are worked under only a single modality setting~\cite{BlockMix_20} due to their high dependence on the similarity of distribution and appearance on either synthetic or real data with neighborhood reference.
Moreover, the designs of the similarity measures~\cite{TANG2022108792} to optimize the cross-modality alignment process have been proven to be quite challenging.
These obstacles gave rise to the development of cross-modality medical image translation~\cite{medireg,LiuLZFL20,Liu_pami_21} and cross-modality Re-ID tasks~\cite{AlignGAN,D2RL}. The former leverages the cGAN~\cite{pix2pix} to perform NeuroImage-to-NeuroImage translation, while the latter learns the cross-modality match in pedestrian images of a person across disjoint camera views by an RGB-to-IR (infrared) translation. A recent study~\cite{NeMAR} presents a multi-modality image registration method by training an image-to-image translation network on the two input modalities.
The basic idea of the above-mentioned approaches is using the image-to-image translation to implement cross-modality transformation to narrow the large variation between different modalities. 

Motivated by this idea, we propose a specialized cross-modality generation-registration paradigm (CGRP) for unsupervised misaligned infrared and visible image fusion.
An inherent characteristic of infrared images is the emphasis on sharp geometry structures over texture details.
%
%
Generating a pseudo infrared image preserving sharp structure information is more conducive to implementing mono-modality registration between it with the corresponding distorted infrared image.
To better preserve geometry structure during generating pseudo infrared images from visible images, we proposed a Cross-modality Perceptual Style Transfer Network (CPSTN), which inherits the basic cycle consistent learning manner of the CycleGAN~\cite{cyclegan}, while developing a perceptual style transfer constraint and cross regularizations across two-cycle learning path to further guide the CPSTN to generate sharper structures. Without them, the generation capability of our CPSTN will degrade to that in~\cite{NeMAR}. 
Doing this lays the foundation for the mono-modality registration of infrared images. We utilize a Multi-level Refinement Registration Network (MRRN) to predict the deformation field from coarse to fine between distorted and pseudo infrared images and reconstruct the registered infrared image. 
In this case, we further perform the fusion of the registered infrared image and visible image.
To enable the fusion network to focus more on faithful texture details, while avoiding feature smoothing caused by some unsophisticated fusion rules (e.g., concatenation, weighted average), we develop an Interaction Fusion Module (IFM) to adaptively select meaningful features from infrared and visible images to achieve fused results with sharp textures.
%
We evaluate the proposed method on manually processed unaligned datasets and demonstrate its strengths through comprehensive analysis. 
The main contributions are summarized as follows:
\begin{itemize}
	\item We propose a highly-robust unsupervised infrared and visible image fusion framework, which is more focused on mitigating ghosting artifacts caused by the fusion of misaligned image pairs compared with learning-based fusion methods specialized for pre-registered images.
	\item Considering the difficulty of the cross-modality image alignment, we exploit a specialized cross-modality generation-registration paradigm to bridge the large discrepancy between modalities, thus achieving effective infrared and visible image alignment. 	
	\item An interaction fusion module is developed to adaptively fuse multi-modality features, which avoids feature smoothing caused by unsophisticated fusion rules and emphasizes faithful texture details.
\end{itemize}
The extensive experimental results demonstrate that the proposed method performs superior capability on misaligned cross-modality image fusion.
The code will be available at~\url{https://github.com/wdhudiekou/UMF-CMGR}.

\begin{figure*}[tbh]
	\centering
	\begin{tabular}{c}
		\includegraphics[width=0.90\linewidth]{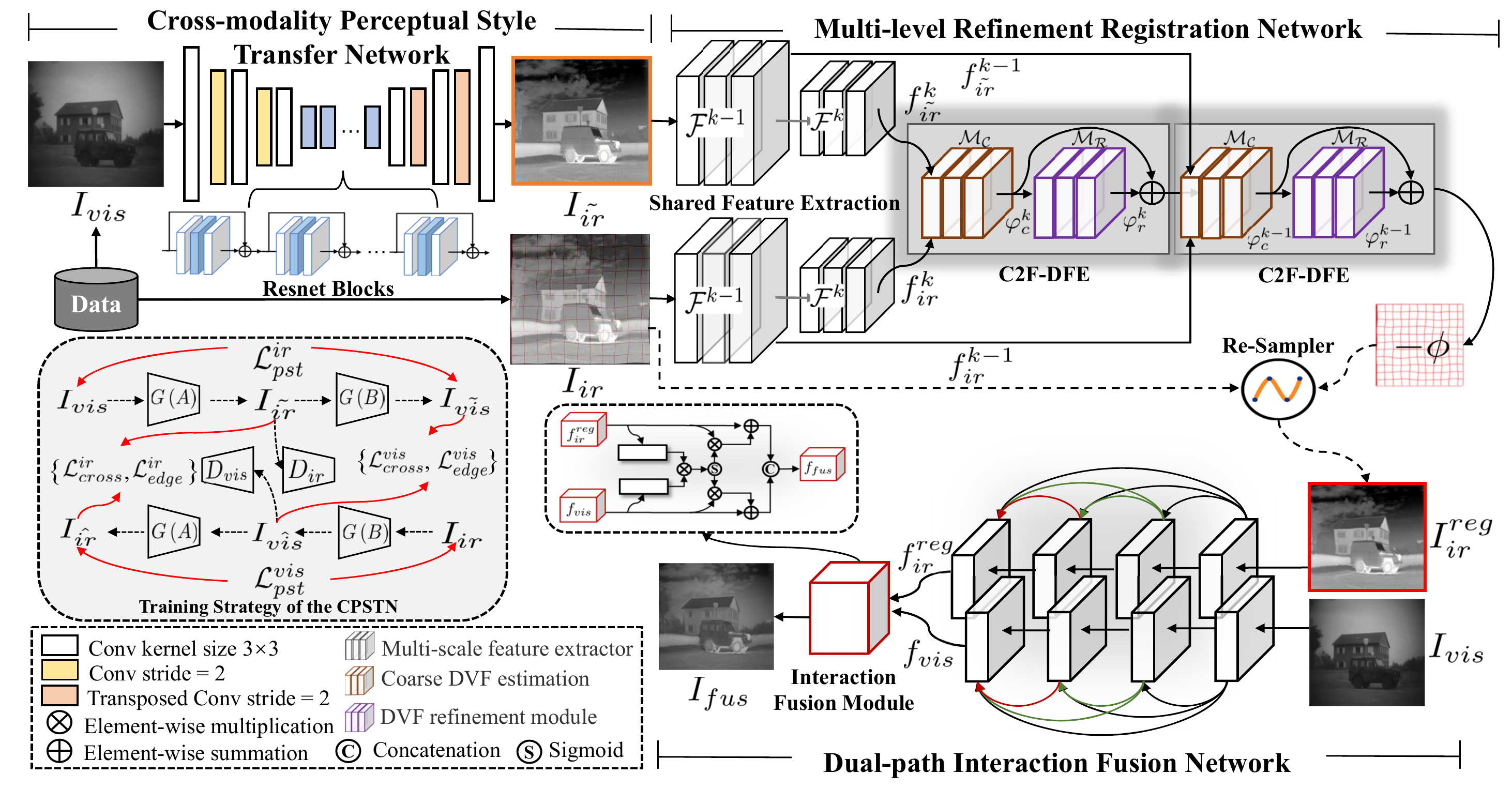} \\
	\end{tabular}
	\vspace{-2mm}
	\caption{The workflow of the proposed unsupervised cross-modality fusion network for misaligned infrared and visible images. The network mainly consists of three subnetworks, which are Cross-modality Perceptual Style Transfer Network, Multi-level Refinement Registration Network, and Dual-path Interaction Fusion Network. Taking misaligned infrared and visible images as inputs, our method performs the above three subnetworks in sequence to obtain the final fused image.}%
	\label{fig:network}
	\vspace{-4mm}
\end{figure*}

\vspace{-2mm}
\section{Method}

\subsection{Motivation}
Different imaging pipelines and heat dissipation within the sensor result in misalignment between observed infrared and visible images, presenting as shift and deformation.
By observing, it is found that straightly fusing misaligned infrared and visible images often suffer serious ghosts.
Inspired by~\cite{AlignGAN,D2RL} that reduce cross-modality variation via image-to-image translation, and catering to the inherent characteristic that infrared image "emphasizes structure over texture", we propose a specialized cross-modality generation-registration paradigm to reduce spatial offsets and alleviate the ghost during misaligned infrared and visible image fusion.
\vspace{-2mm}
\subsection{Cross-modality Perceptual Style Transfer}
The first part of the proposed cross-modality generation-registration paradigm (CGRP) is image translation.
Considering that infrared images are prone to distortion due to thermal radiation, we proposed a Cross-modality Perceptual Style Transfer Network (CPSTN) to translate the visible image $I_{vis}$ to its infrared counterpart $I_{\tilde{ir}}$. Together, a pseudo infrared image pair $\left(I_{ir}, I_{\tilde{ir}}\right)$ is formed to provide a unified representation.
%
As shown in Figure~\ref{fig:network}, CPSTN is a UNet-like generator, and its bottom obtains $9$ resnet blocks.
Different from CycleGAN~\cite{cyclegan}, we tend to design a specific learning strategy controlled by a perceptual style transfer constraint and establish the inter-path correlation between the two-cycle generative routes to further optimize~\cite{liu2022general,liu2021investigating} the generation of sharp structure of the pseudo infrared image. The pseudo infrared image is generated by $I_{\tilde{ir}}=\mathcal{T}_{\theta}(I_{ir})$, where $\mathcal{T}_{\theta}$ denotes the CSPTN with network parameter $\theta$.
This optimization process is illustrated in the left dotted box in Figure~\ref{fig:network}. Note that the $G(A)$ corresponds to our CPSTN, and the perceptual style transfer constraint and the regularization of inter-path correlation are explained in Section~\ref{subsec:loss}.

\subsection{Multi-level Refinement Registration}
Thanks to the reduction of cross-modality discrepancy by CPSTN, the registration of infrared images under the mono-modality settings becomes the other core part of the CGRP.
As shown in Figure~\ref{fig:network}, we exploit a Multi-level Refinement Registration Network (MMRN) to predict deformation field between distorted and pseudo infrared images and reconstruct the registered infrared image.
The MRRN consists of a shared multi-level feature extractor (SM-FE), two Coarse-to-Fine deformation field estimation (C2F-DFE) modules, and a resampler layer. In each C2F-DFE, a coarse DFE module $\mathcal{M}_{\mathcal{C}}$ and a refined DFE module $\mathcal{M}_{\mathcal{R}}$ are included.
Then, the coarse deformation field is first predicted as
\begin{small}
	\begin{equation}
	\vspace{-1.5mm}
		\begin{split}
			\varphi_{c}^{k}=\mathcal{M}_{\mathcal{C}}\left(\mathcal{F}^{k}(I_{\tilde{ir}},I_{ir}) \right),
		\end{split}
		\vspace{-1.5mm}
	\end{equation}
\end{small}
\vspace{-1mm}
the refined deformation field is estimated by
\begin{small}
	\begin{equation}
	\vspace{-1.5mm}
		\begin{split}
			\varphi_{r}^{k}=\mathcal{M}_{\mathcal{R}}\left(\varphi_{c}^{k} \right) \oplus \varphi_{c}^{k},
		\end{split}
		\vspace{-1.5mm}
	\end{equation}
\end{small}
\hspace{-1mm}where $\mathcal{F}^{k}$ denotes the $k$-th level of the SM-FM. Suppose the SM-FE contains $K$ levels, when $k=K$, the final deformation field $-\phi=\varphi_{r}^{K}$ is estimated.
Finally, we use the resampler layer similar to the STN~\cite{STN} to reconstruct the registered infrared image by
\begin{small}
	\begin{equation}
	\vspace{-1.5mm}
		\begin{split}
			I_{ir}^{reg} = I_{ir} \circ (-\phi),
		\end{split}
	\vspace{-1.5mm}
	\end{equation}
\end{small}
\hspace{-1mm}the operator $\circ$ denotes the spatial transform for registration.

\subsection{Dual-path Interation Fusion}
To fuse registered infrared image $I_{ir}^{reg}$ and visible image $I_{vis}$, we present a dual-path interaction fusion network (DIFN). It is composed of a dual-path feature extraction module and a feature interaction fusion module (IFM). Here, the structure of the dual-path feature extraction module inherits the residual dense network~\cite{RDN}, and the extracted features are presented by
\begin{small}
	\begin{equation}
		f_{ir}^{reg},f_{vis}=\mathcal{M}_{\theta_{E}}\left(I_{ir}^{reg},I_{vis}\right), 
	\end{equation}
\end{small}
\hspace{-1mm}where $\mathcal{M}_{\theta_{E}}$ is the feature extraction module, and $\theta_{E}$ is its parameter.

\noindent
\textbf{Interaction Fusion Module.}
We utilize the IFM to adaptively select features from infrared and visible images to fuse, shown in small dashed box of Figure~\ref{fig:network}. To focus on more significant information, the feature responses are recalibrated by
\vspace{-3.0mm}
\begin{small}
	\begin{equation}
	\vspace{-1.5mm}
		\begin{split}
			\mathcal{A}_{tt}=\mathcal{S}\left( Conv_{1\times1}\left(f_{ir}^{reg}\right) \otimes Conv_{1\times1}\left(f_{vis}\right) \right),
		\end{split}
	\vspace{-1.5mm}
	\end{equation}
\end{small}
\hspace{-1mm}and the infrared and visible features are activated as
\begin{small}
	\begin{equation}
	\vspace{-1.0mm}
		\begin{split}
			f_{ir}^{Att} = f_{ir}^{reg}\otimes\left(1+\mathcal{A}_{tt}\right),\\
			f_{vis}^{Att} = f_{vis}\otimes\left(1+\mathcal{A}_{tt}\right),
		\end{split}
	\vspace{-1.0mm}
	\end{equation}
\end{small}
\hspace{-1mm}then we obtain the final fused image by
\begin{small}
	\begin{equation}
	\vspace{-1.5mm}
		\begin{split}
			I_{fus} = Conv_{3 \times 3}\left(Concat\left(f_{ir}^{Att},f_{vi}^{Att}\right)\right),
		\end{split}
	\vspace{-1.5mm}
	\end{equation}
\end{small}
\hspace{-1mm}where $\mathcal{S}$ is the Sigmoid function, and $\otimes$ denotes the element-wise multiplication operation.
\vspace{-1mm}

\subsection{Loss Functions}\label{subsec:loss}
\noindent
\textbf{Perceptual style transfer loss.}
To generate more realistic pseudo infrared images, we introduce a perceptual style transfer (PST) loss to control the cycle consistency of CPSTN. The PST loss consists of two terms known as perceptual loss $\mathcal{L}_{pcp}$ and style loss $\mathcal{L}_{sty}$. First, the $\mathcal{L}_{pcp}$ is defined as
\begin{small}
	\begin{equation}
		\vspace{-1.0mm}
		\begin{split}
			\mathcal{L}_{pcp}^{\psi_{j}}=
			\|\psi_{j}(I_{vis})-\psi_{j}\left(G_{B}\left(G_{A}(I_{vis})\right)\right) \|^{2} \\ + \|\psi_{j}(I_{ir})-\psi_{j}\left(G_{A}\left(G_{B}(I_{ir})\right)\right) \|^{2},
		\end{split}
		\vspace{-1.0mm}
	\end{equation}
\end{small}
\hspace{-1.3mm}where $\psi_{j}$ is the $j$-th layer of the VGG-$19$~\cite{vgg} model and $j\in[2,7,12,21,30]$ along with weight $\omega \in[\frac{1}{32},\frac{1}{16},\frac{1}{8},1,1]$.
These features are also used to compute the $\mathcal{L}_{sty}$, which is defined as
\vspace{-1.0mm}
\begin{small}
	\begin{equation}
		\begin{split}
			\mathcal{L}_{sty}^{\psi_{j}}=\omega_{j}\| \mathcal{G}_{\psi_{j}}(I_{vis})-\mathcal{G}_{\psi_{j}}\left(G_{B}\left(G_{A}(I_{vis})\right)\right) \|^{2}\\ + \omega_{j}\| \mathcal{G}_{\psi_{j}}(I_{ir})-\mathcal{G}_{\psi_{j}}\left(G_{A}\left(G_{B}(I_{ir})\right)\right) \|^{2},
		\end{split}
	\end{equation}
\end{small}
\hspace{-1.5mm}where $\mathcal{G}$ is Gram matrix~\cite{GramMatrix}, used to combat “checkerboard” artifacts. The overall PST loss is
\begin{small}
	\begin{equation}
		\mathcal{L}_{pst}=\lambda_{p}\mathcal{L}_{pcp} + \lambda_{s}\mathcal{L}_{sty},
	\end{equation}
\end{small}
\hspace{-1.0mm}where $\lambda_{p}$ is set to $1.0$, $\lambda_{s}$ is set to $100$.
\vspace{1.0mm}

\noindent
\textbf{Cross regularization loss.}
Inventively, we present a cross regularization between two-cycle paths during training CPSTN to establish the inter-path correlation. It contains both content term $\mathcal{L}_{con}$ and edge term $\mathcal{L}_{edge}$, which are defined as
\vspace{-2.0mm}
\begin{small}
	\begin{equation}
		\begin{split}
			\label{eq: cross-loss}
			\vspace{-2.5mm}
			\mathcal{L}_{con}&=\|I_{\tilde{ir}}-I_{\hat{ir}}\|_{1}+\|I_{v\tilde{i}s}-I_{v\hat{i}s}\|_{1}, \\
			\mathcal{L}_{edge}&=\|\nabla I_{\tilde{ir}}-\nabla I_{\hat{ir}}\|_{char}+\|\nabla I_{v\tilde{i}s}-\nabla I_{v\hat{i}s}\|_{char},	
		\end{split}
	\end{equation}
\end{small}
\hspace{-1.0mm}where $\nabla$ denotes laplacian gradient operator. We compute $\mathcal{L}_{edge}$ using the Charbonnier Loss~\cite{CharLoss}. The overall cross regularization is computed by
\begin{small}
	\begin{equation}
		\mathcal{L}_{cross}=\lambda_{c}\mathcal{L}_{con} + \lambda_{e}\mathcal{L}_{edge},
	\end{equation}
\end{small}
\hspace{-1.0mm}where, $\lambda_{c}=\lambda_{e}=8.0$.
\vspace{1.0mm}

\noindent
\textbf{Registration loss.}
We adopt a bidirectional similarity loss to constrain the registration between distorted and pseudo infrared images in feature space, which is defined as
\begin{small}
	\begin{equation}
		\mathcal{L}_{sim}^{bi}=\|\psi_{j}(I_{ir}^{reg})-\psi_{j}(I_{\tilde{ir}}) \| + \lambda_{rev} \|\psi_{j}(\phi \circ I_{\tilde{ir}})-\psi_{j}(I_{ir})\|_{1},
	\end{equation}
\end{small}
\hspace{-1.0mm}the first term is forward, while the second term is backward with weight $\lambda_{rev}=0.2$, in which the reverse deformation field $\phi$ is used to distort the pseudo infrared image $I_{\tilde{ir}}$ and make it approach to the distorted input $I_{ir}$.
To ensure a smooth deformation field, we define a smooth loss as
\begin{small}
	\begin{equation}
		\mathcal{L}_{\text {smooth }}=\|\nabla \phi\|_{1}.
	\end{equation}
\end{small}
%
And then, the overall registration loss is computed by
\begin{equation}
	\mathcal{L}_{\mathrm{reg}}=\mathcal{L}_{sim}+\lambda_{sm} \mathcal{L}_{smooth},
\end{equation}
where $\lambda_{sm}$ is set to $10$ in our work.
\vspace{1.0mm}

\noindent
\textbf{Fusion loss.}
In the fuison phase, we utilize MS-SSIM loss function $\mathcal{L}_{ssim}^{ms}$ to maintain sharper intensity distribution of the fused image, which is defined as
\begin{small}
	\begin{equation}
		\mathcal{L}_{ssim}^{ms}=\left(1-SSIM(I_{fus}, I_{ir}^{reg})\right)+\left(1-SSIM(I_{fus}, I_{vis})\right).
	\end{equation}
\end{small}
\hspace{-1mm}To encourage the restoration of texture details, we model the gradient distribution and develop a joint gradient loss as
\begin{small}
\begin{equation}
	\mathcal{L}_{JGrad}=\|max\left(\nabla I_{ir}^{reg},\nabla I_{vis}\right), \nabla I_{fus} \|_{1}.
\end{equation}
\end{small}
\hspace{-1.0mm}For clearer textures, the gradient of the fused image is forced to approach the maximum value between the infrared and visible image gradients. 
In addition, to retain the saliency targets from the two images, we leverage self-visual saliency maps inspried by~\cite{SvSW,SMoA_21,TarDAL_22} to build a new loss as 
\begin{small}
\begin{equation}
	\begin{split}
		\boldsymbol{\omega}_{ir}&=\boldsymbol{S}_{f_{i r}} /\left(\boldsymbol{S}_{f_{i r}}-\boldsymbol{S}_{f_{v i s}}\right),
		\boldsymbol{\omega}_{vis}=1-\boldsymbol{\omega}_{ir}, \\
		\mathcal{L}_{svs}&=\|\left(\boldsymbol{\omega}_{ir} \otimes I_{ir}^{reg}+\boldsymbol{\omega}_{vis} \otimes I_{vis}\right), I_{fus} \|_{1},
	\end{split}
\end{equation}
\end{small}
\hspace{-1.3mm}where $\boldsymbol{S}$ denotes saliency matrix, $\boldsymbol{\omega}_{vis}$ and $\boldsymbol{\omega}_{ir}$ denote the weight maps for infrared and visible images, respectively.
The overall fusion loss is computed by
\begin{equation}
	\mathcal{L}_{fus}=\lambda_{ssim}\mathcal{L}_{ssim}^{ms} + \lambda_{JG}\mathcal{L}_{JGrad} + \lambda_{svs}\mathcal{L}_{svs},
\end{equation}
where $\lambda_{ssim}$, $\lambda_{JG}$, and $\lambda_{svs}$ are set to $1.0$, $20.0$, and $5.0$.

\noindent
\textbf{Overall loss.}
We train our network by minimizing the following overall loss function
\begin{equation}
	\mathcal{L}_{total}=\mathcal{L}_{pst} +\mathcal{L}_{cross} + \mathcal{L}_{GAN} + \mathcal{L}_{reg} + \mathcal{L}_{fus}.
\end{equation}
Note that the $\mathcal{L}_{GAN}$ inherits that of CycleGAN to discriminate whether pseudo infrared image is real or fake. 

\section{Experiments}

\subsection{Dataset and Implement Details}

\textbf{Datasets.} Our misalignment cross-modality image fusion are conducted on two widely-used datasets: TNO\footnote{http://figshare.com/articles/TNO\_Image\_Fusion\_Dataset/1008029.}, and RoadScene\footnote{https://github.com/hanna-xu/RoadScene.}, which are pre-registered.
To meet the requirement of misaligned image pairs, we first generate several deformation fields by performing different degrees of affine and elastic transformations and then apply them to infrared images to obtain distorted images.
%
We use all $221$ images from RoadScene dataset as training samples. Due to the limited amount of the TNO dataset, we only use it as a testing set. In addition, we randomly select $55\%$ images from RoadScene dataset for testing since our method is full-unsupervised.   

\noindent
\textbf{Implement Details.}
The code of our method is implemented using the PyTorch framework with an NVIDIA 2080Ti GPU.
We randomly select $8$ image patches of size $256\times256$ to form a batch. The Adam optimizer ($\beta _{1}=0.9$, and $\beta _{2}=0.999$) is used to optimize our model. The initial learning rate is set to $0.001$ and remains unchanged throughout the training phase that goes through $300$ epochs.

\subsection{Evaluation in IR-VIS Image Registration}
We evaluate the middle registered results using three common metrics including MSE, Mutual Information (MI)~\cite{MI_metric}, and Normalized Cross Correlation (NCC)~\cite{NCC_metric}.
Considering that multi-modality image alignment methods are extremely rare, for a fair comparison with ours, we choose two typical image alignment algorithms based on deformation field, which are FlowNet~\cite{flownet} and VoxelMorph~\cite{voxelmorph}. Note that, FlowNet only predicts the deformation field for input image pairs, so the spatial transform layer (STN)~\cite{STN} is employed to generate registered images.  
As reported by Table~\ref{tab:align-resluts}, using directly these two algorithms to perform cross-modality alignment of the IR-VIS images yields negligible improvement, even worse than the misaligned inputs.
In contrast, the proposed CGRP gains about $\textbf{0.10}$ and $\textbf{0.23}$ improvments in MI on TNO and RoadScene datasets, respectively.
It is demonstrated that the proposed CGRP is more effective than existing popular image alignment methods.

\begin{table}[h]
	\scriptsize
	\begin{center}
		\vspace{-0.2cm}
		\renewcommand\arraystretch{1.2} 
		\setlength{\tabcolsep}{4.6pt}{ 
			\begin{tabular}{lcccccc}
				\toprule
				\multirow{2}{4em}{\textbf{Methods}}&\multicolumn{3}{c}{\textbf{TNO}} &\multicolumn{3}{c}{\textbf{RoadScene}}\\
				& MSE$\downarrow$ & NCC$\uparrow$ & MI$\uparrow$ & MSE$\downarrow$ & NCC$\uparrow$ & MI$\uparrow$\\
				\midrule
				\specialrule{0em}{1pt}{1pt}
				Misaligned Input & $0.007$ & $0.876$ & $1.558$  & $0.01$ & $0.894$ & $1.602$ \\
				\cdashline{1-7}[0.8pt/2pt]
				\specialrule{0em}{1pt}{1pt}
				FlowNet + STN & $0.029$ & $0.636$ & $1.095$ & $0.009$ & $0.910$ & $1.549$ \\
				FlowNet + STN + CPST & $0.007$ & $0.893$ & $1.465$ & $0.005$ & $0.949$ & $1.744$ \\
				\cdashline{1-7}[0.8pt/2pt]
				\specialrule{0em}{1pt}{1pt}
				VoxelMorph & $0.007$ & $0.880$ & $1.545$ & $0.008$ & $0.914$ & $1.589$ \\
				VoxelMorph + CPST & $0.005$ & $0.919$ & $1.573$ & $0.006$ & $0.941$ & $1.689$ \\
				\cdashline{1-7}[0.8pt/2pt]
				\specialrule{0em}{1pt}{1pt}
				Ours CGRP & $\textcolor{red}{\textbf{0.004}}$ & $\textcolor{red}{\textbf{0.926}}$ & $\textcolor{red}{\textbf{1.648}}$ & $\textcolor{red}{\textbf{0.004}}$ & $\textcolor{red}{\textbf{0.963}}$ & $\textcolor{red}{\textbf{1.833}}$ \\
				\bottomrule
		\end{tabular}}
		\vspace{-2mm}
		\caption{Quantitative comparisons of cross-modality image registration on the TNO and Roadscene datasets. The best results are presented with \textbf{\textcolor{red}{red}} bold font.}
		\label{tab:align-resluts}
	\end{center}
	\vspace{-8mm}
\end{table}

\subsection{Comparing with SOTA in IVIF}

We perform quantitative and qualitative evaluations for the proposed method against several state-of-the-art IVIF methods, including DENSE~\cite{DenseFuse}, DIDFuse~\cite{DIDFuse_2020}, FGAN~\cite{FGAN}, GAN-FM~\cite{GAN-FM}, MFEIF~\cite{MFEIF}, RFN~\cite{RFN}, and U2F~\cite{U2Fusion}, equipped with FlowNet+STN and VoxelMorph, two typical image alignment algorithms used to cross-modality image registration in our work.
For fair comparisons, we use the same testing samples ($121$ and $24$ images from RoadScene and TNO datasets) to evaluate them. 

\noindent
\textbf{Quantitative evaluation.}
We report the quantitative fusion results on TNO and RoadScene datasets in Table~\ref{tab:quanti-results}.
It can be seen that our methods numerically outperforms existing IVIF methods by large margins and ranks first across all three metrics, including Cross Correlation (CC), Visual Information Fidelity (VIF)~\cite{vif}, and Structural Similarity Index (SSIM)~\cite{ssim}.
Especially, compared with GAN-FM, a recent research, our method
gains $\textbf{0.33}$ and $\textbf{0.46}$ improvements in VIF on TNO and RoadScene datasets, which demonstrate the superiority of the proposed method.

\begin{table*}[h]
\footnotesize
	\begin{center}
		\centering
		\setlength{\tabcolsep}{5.0pt}{
			\begin{tabular}{cccccccccc}
				\toprule
				& DENSE & DIDFuse & FGAN & GAN-FM & MFEIF & PMGI & RFN & U2F & Ours \\
				\midrule
				\specialrule{0em}{1pt}{1pt}
				& \multicolumn{9}{l}{\ding{172}: \textbf{Aligned Method: (FlowNet+STN) / VoxelMorph} \ \ \ \ \ding{173}: \textbf{Dataset: RoadScene} }\\
				\hline
				\specialrule{0em}{1pt}{1pt}
				CC$\uparrow$& $0.589$/$\textcolor{blue}{\textbf{0.597}}$   & $0.573$/$0.582$ & $0.502$/$0.582$  & $0.565$/$0.575$  & $0.588$/$0.596$  &  $0.530$/$0.538$  & $0.587$/$0.593$  & $0.558$/$0.569$ & $\textcolor{red}{\textbf{0.621}}$  \\
				\specialrule{0em}{1pt}{1pt} 
				VIF$\uparrow$& $0.488$/$0.521$  & $0.445$/$0.469$ & $0.493$/$0.513$  & $0.564$/$0.597$  & $0.619$/$\textcolor{blue}{\textbf{0.659}}$   &  $0.408$/$0.447$  & $0.563$/$0.590$  & $0.430$/$0.464$ & $\textcolor{red}{\textbf{0.895}}$ \\
				\specialrule{0em}{1pt}{1pt}
				SSIM$\uparrow$& $0.311$/$0.355$ & $0.301$/$0.335$ & $0.255$/$0.281$  & $0.334$/$0.368$  & $0.342$/$\textcolor{blue}{\textbf{0.380}}$   &  $0.266$/$0.314$  & $0.319$/$0.347$  & $0.297$/$0.343$ & $\textcolor{red}{\textbf{0.507}}$ \\
				\hline
				\specialrule{0em}{1pt}{1pt}	
				& \multicolumn{9}{l}{\ding{172}: \textbf{{Aligned Method: (FlowNet+STN) / VoxelMorph} }\ \ \ \ \ding{173}: \textbf{Dataset: TNO} }\\
				\hline
				\specialrule{0em}{1pt}{1pt}
				CC$\uparrow$& $0.390$/$0.463$ & $0.380$/$0.452$  & $0.315$/$0.400$ & $0.350$/$0.419$ & $0.381$/$0.451$  & $0.363$/$0.447$ & $0.392$/$\textcolor{blue}{\textbf{0.465}}$ & $0.378$/$0.456$ & $\textcolor{red}{\textbf{0.481}}$ \\
				\specialrule{0em}{1pt}{1pt}
				VIF$\uparrow$& $0.675$/$0.742$ & $0.545$/$0.597$  & $0.588$/$0.654$  & $0.556$/$0.627$  & $0.701$/$\textcolor{blue}{\textbf{0.780}}$ & $0.518$/$0.606$ & $0.671$/$0.728$ & $0.535$/$0.601$ & $\textcolor{red}{\textbf{1.016}}$ \\
				\specialrule{0em}{1pt}{1pt}
				SSIM& $0.318$/$\textcolor{blue}{\textbf{0.382}}$ & $0.266$/$0.328$  & $0.236$/$0.294$  & $0.295$/$0.372$  & $0.297$/$0.379$   &  $0.277$/$0.363$  & $0.300$/$0.349$  & $0.299$/$0.372$ & $\textcolor{red}{\textbf{0.473}}$  \\
				\specialrule{0em}{1pt}{1pt}	
				\bottomrule
		\end{tabular}}
	\end{center}
	\vspace{-4mm}  
	\caption{The quantitative comparisons of the IVIF results with eight state-of-the-art fusion methods on two common datasets when using FlowNet+STN and VoxelMorph algorithms as the basic registration models.}
	\label{tab:quanti-results}
	\vspace{-2mm}
	\footnotesize
\end{table*}

\begin{figure*}[t]
	\begin{center}
		\begin{tabular}{cccccccccc}	
			
			\includegraphics[width = 0.153\linewidth,height=0.10\textheight]{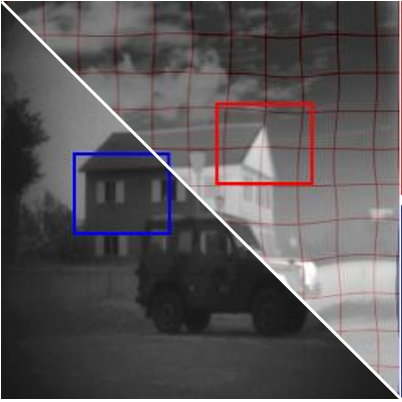} & 
			\hspace{-0.46cm}
			\includegraphics[width = 0.09\linewidth,height=0.10\textheight]{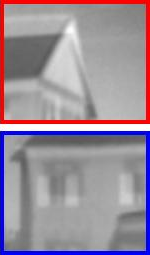}  & 
			\hspace{-0.46cm}
			\includegraphics[width = 0.09\linewidth,height=0.10\textheight]{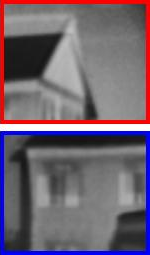}& 
			\hspace{-0.46cm}
			\includegraphics[width = 0.09\linewidth,height=0.10\textheight]{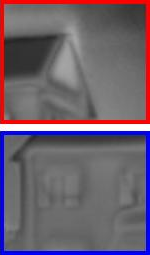} & 
			\hspace{-0.46cm}
			\includegraphics[width = 0.09\linewidth,height=0.10\textheight]{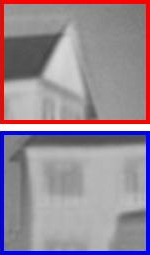} & 
			\hspace{-0.46cm}
			\includegraphics[width = 0.09\linewidth,height=0.10\textheight]{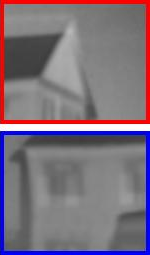} &
			\hspace{-0.46cm}
			\includegraphics[width = 0.09\linewidth,height=0.10\textheight]{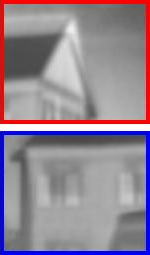} &
			\hspace{-0.46cm}
			\includegraphics[width = 0.09\linewidth,height=0.10\textheight]{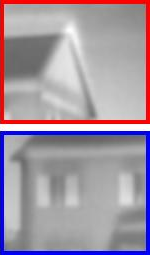} &
			\hspace{-0.46cm}
			\includegraphics[width = 0.09\linewidth,height=0.10\textheight]{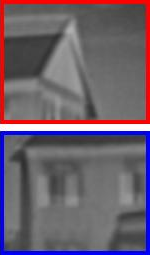} &
			\hspace{-0.46cm}
			\includegraphics[width = 0.09\linewidth,height=0.10\textheight]{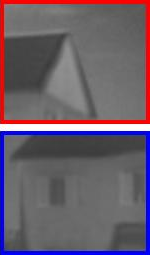} \\

			\includegraphics[width = 0.15\linewidth,height=0.10\textheight]{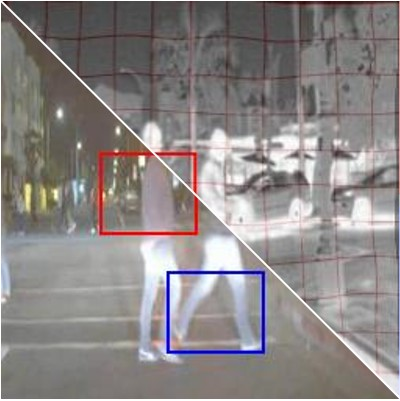}  & \hspace{-0.46cm}
			\includegraphics[width = 0.09\linewidth,height=0.10\textheight]{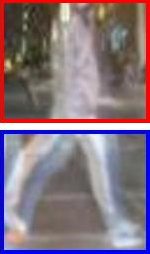}  & \hspace{-0.46cm}
			\includegraphics[width = 0.09\linewidth,height=0.10\textheight]{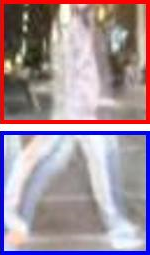} & \hspace{-0.46cm}
			\includegraphics[width = 0.09\linewidth,height=0.10\textheight]{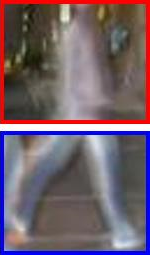} & \hspace{-0.46cm}
			\includegraphics[width = 0.0895\linewidth,height=0.10\textheight]{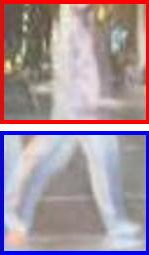} & \hspace{-0.46cm}
			
			\includegraphics[width = 0.09\linewidth,height=0.10\textheight]{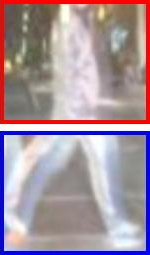} &
			\hspace{-0.46cm}
			\includegraphics[width = 0.09\linewidth,height=0.10\textheight]{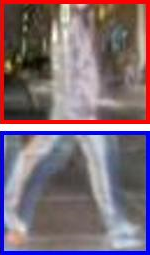} &
			\hspace{-0.46cm}
			\includegraphics[width = 0.09\linewidth,height=0.10\textheight]{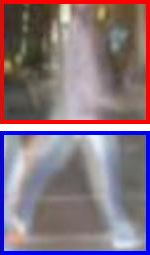} &
			\hspace{-0.46cm}
			\includegraphics[width = 0.09\linewidth,height=0.10\textheight]{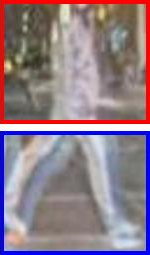} &
			\hspace{-0.46cm}
			\includegraphics[width = 0.09\linewidth,height=0.10\textheight]{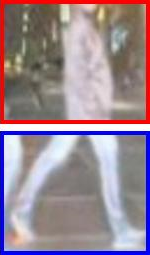} \\		
			
			\small VIS/IR
			& \hspace{-0.46cm} \small DENSE
			& \hspace{-0.46cm} \small DIDFuse
			& \hspace{-0.46cm} \small FGAN
			& \hspace{-0.46cm} \small GAN-FM 
						
			& \hspace{-0.46cm} \small MFEIF
			& \hspace{-0.46cm} \small PMGI
			& \hspace{-0.46cm} \small RFN
			& \hspace{-0.46cm} \small U2F
			& \hspace{-0.46cm} \small Ours			
			\\
		\end{tabular}
	\end{center}
	\vspace{-4mm}
	\caption{Visualization of the IVIF results from different methods on the TNO and RoadScene datasets when using the FlowNet+STN algorithm as the basic registration model.}
	\label{fig:flownet_tno}
	\vspace{-4mm}
\end{figure*}

\noindent
\textbf{Qualitative evaluation.}
%
Due to space limitation, Figure~\ref{fig:flownet_tno} only presents qualitative comparisons using FlowNet+STN as the basic registration model.
By observing the local enlarged areas, we can find that prior works fail to achieve desirable alignment and ghost elimination. In contrast, the proposed method shows favorable alignment and fusion capability while preserving sharp structures.


\begin{table}[t]
	\scriptsize
	\begin{center}
		\vspace{-0.2cm}
		\renewcommand\arraystretch{1.3} 
		\setlength{\tabcolsep}{1.8pt}{ 
			\begin{tabular}{lccccccccc}
				\toprule
				& DENSE & FGAN & DIDFuse & U2F & PMGI & RFN & MFEIF & GAN-FM & Ours\\
				
				\midrule
				\specialrule{0em}{1pt}{1pt}
				\textbf{Param.} & $0.925$  & $\textbf{\textcolor{blue}{0.074}}$ & $0.261$ & $0.659$  & $\textbf{\textcolor{red}{0.042}}$ & $10.94$ & $0.158$ & $10.21$ &$0.80$\\
				\textbf{Time} & $0.124$  & $0.251$ & $0.055$ & $0.123$  & $0.182$ & $0.239$ & $\textbf{\textcolor{blue}{0.045}}$ & $0.334$ & $\textbf{\textcolor{red}{0.024}}$\\
				
				\bottomrule
		\end{tabular}}
		\vspace{-2mm}
		\caption{Efficiency analysis of our joint registration-fusion model against state-of-the-art fusion methods.}
		\label{tab:efficiency}
	\end{center}
	\vspace{-6mm}
\end{table}
\noindent
\textbf{Model efficiency.}
To examine model efficiency, we report the statistics of Params(M) and RunTime(s) in Table~\ref{tab:efficiency}. We conduct the measures for a paired images with size of $64\times64$ on a single 2080Ti GPU. Noted that, our results are obtained by jointly measuring registration and fusion networks, while other methods only have a single fusion network. The results show that our method is competitive at least in running time.

\subsection{Ablation Studies}\label{subs:ab}

\begin{figure}[t]
	\begin{center}
		\begin{tabular}{ccc}
			\hspace{-0.20cm}
			\includegraphics[width = 0.33\linewidth]{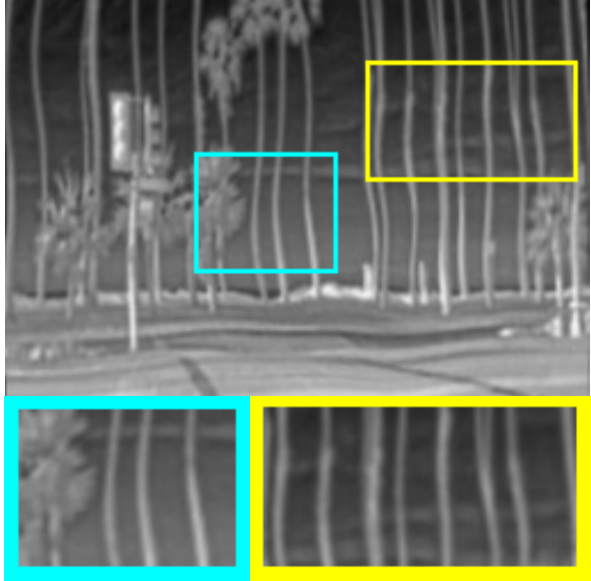}   & \hspace{-0.46cm}
			\includegraphics[width = 0.33\linewidth]{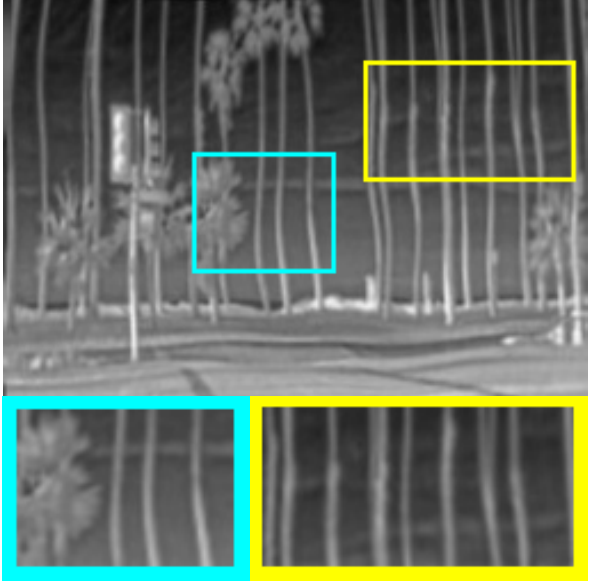} & \hspace{-0.46cm}
			\includegraphics[width = 0.33\linewidth]{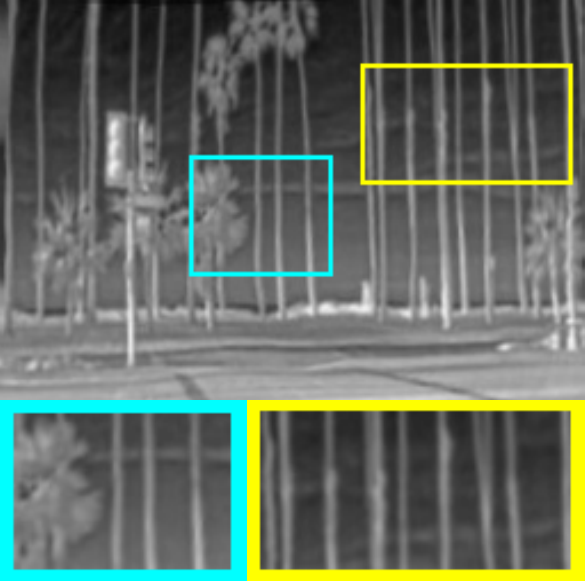} \\		
			{\small (a) FlowNet}
			& \hspace{-0.36cm} {\small (b) FlowNet+CPSTN}
			& \hspace{-0.36cm} {\small (c) Ours}\\			
			
		\end{tabular}
	\end{center}
	\vspace{-4mm}
	\caption{Ablation analysis of the CPSTN from the registration perspective on RoadScene dataset.}
	\label{fig:fig_ab_CMIG}
	\vspace{-4mm}
\end{figure}

\begin{figure}[t]
	\begin{center}
		\begin{tabular}{ccc}
			\hspace{-0.20cm}
			\includegraphics[width = 0.33\linewidth]{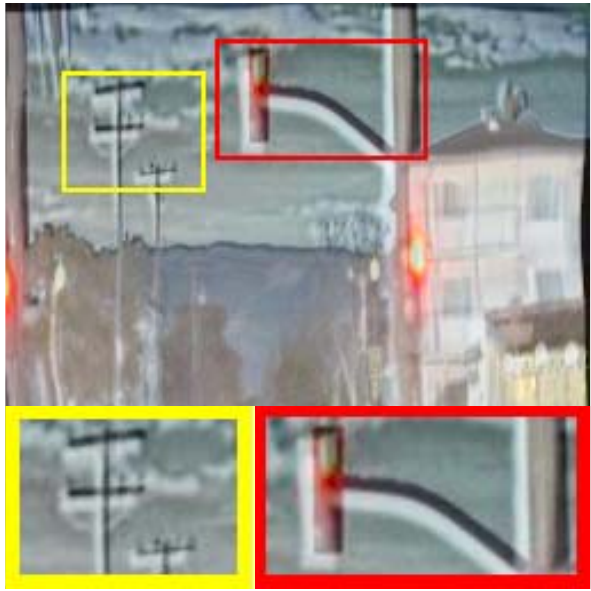}   & \hspace{-0.46cm}
			\includegraphics[width = 0.325\linewidth]{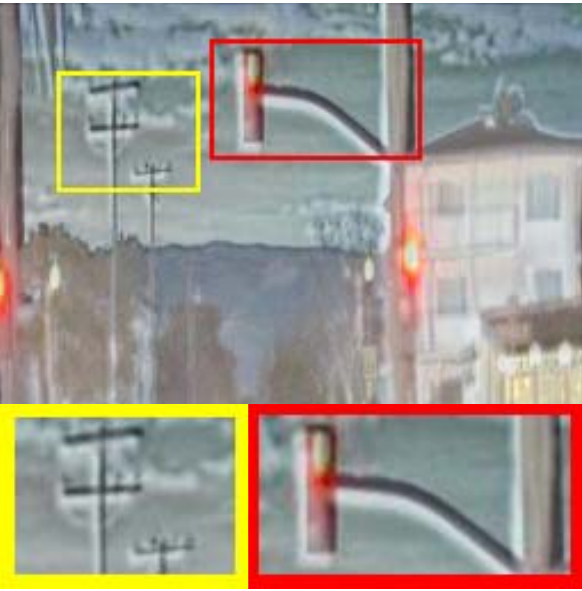} & \hspace{-0.46cm}
			\includegraphics[width = 0.331\linewidth]{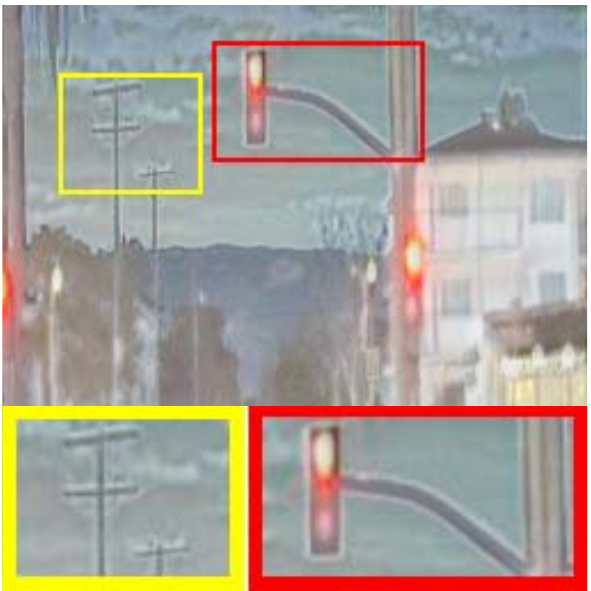} \\						
			
			{\small (a) w/o $\mathcal{T}$,$\mathcal{R}$}
			& \hspace{-0.36cm} {\small (b) w/o $\mathcal{T}$}
			& \hspace{-0.36cm} {\small (c) Ours}\\				
			
		\end{tabular}
	\end{center}
	\vspace{-4mm}
	\caption{Ablation analysis of the CPSTN ($\mathcal{T}$) and MRRN ($\mathcal{R}$) on RoadScene dataset.}
	\label{fig:our_ablation}
	\vspace{-4mm}
\end{figure}

\noindent
\textbf{CPSTN.}
We plug the CPSTN into FlowNet and VoxelMorph as their improved versions, then adopt them to conduct cross-modality IR-VIS image registration. As shown in Table~\ref{tab:align-resluts}, their quantitative results improve a large margin in collaboration with CPSTN than the original versions. Accordingly, the visual comparisons provided in Figure~\ref{fig:fig_ab_CMIG} suggest the effectiveness of the CPSTN. We observe that the registered results generated by the FlowNet model equipped with the CPSTN eliminate evident distortion.
Besides, we verify the effectiveness of the CPSTN from the perspective of image fusion. As quantified in Table~\ref{tab:our_ablation}, improvements of $\textbf{0.14}$ and $\textbf{0.22}$ in VIF are obtained using the CPSTN on TNO and RoadScene datasets, respectively. The corresponding qualitative comparisons in Figure~\ref{fig:our_ablation}(b) and (c) show that the CPSTN contributes to favorable fusion results with negligible ghosts for misaligned infrared and visible images.
The above results comprehensively reveal the effectiveness of CPSTN from registration and fusion perspectives.
\vspace{1.0mm}

\begin{table}[t]
	\scriptsize
	\begin{center}
		\vspace{-0.2cm}
		\renewcommand\arraystretch{1.2} 
		\setlength{\tabcolsep}{2.8pt}{ 
			\begin{tabular}{ccccccc}
				\toprule
				\multirow{2}{3.3em}{\textbf{Datasets}}&\multirow{2}{3.3em}{\textbf{CPSTN}}&\multirow{2}{3.3em}{\textbf{MRRN}}&\multirow{2}{3.3em}{\textbf{DIFN}}&\multicolumn{3}{c}{\textbf{Metrics}} \\
				&& & &CC$\uparrow$ & SSIM$\uparrow$ & VIF$\uparrow$\\
				\midrule
				\specialrule{0em}{1pt}{1pt}
				&\xmark & \xmark & \cmark & $0.438$ & $0.369$ & $0.732$ \\
				TNO&\xmark & \cmark & \cmark & $0.476_{\color{red}{\uparrow{0.038}}}$ & $0.418_{\color{red}{\uparrow{0.049}}}$ & $0.876_{\color{red}{\uparrow{0.144}}}$ \\
				
				\cdashline{2-4}[0.8pt/2pt]\cdashline{5-7}[0.8pt/2pt]
				\specialrule{0em}{1pt}{1pt}
				&\cmark & \cmark & \cmark & $\textbf{0.481}_{\color{red}{\uparrow{0.044}}}$ & $\textbf{0.473}_{\color{red}{\uparrow{0.103}}}$ & $\textbf{1.016}_{\color{red}{\uparrow{0.374}}}$ \\
				\midrule
				\specialrule{0em}{1pt}{1pt}
				&\xmark & \xmark & \cmark & $0.563$ & $0.367$ & $0.543$ \\
				Road&\xmark & \cmark & \cmark & $0.601_{\color{red}{\uparrow{0.038}}}$ & $0.422_{\color{red}{\uparrow{0.055}}}$ & $0.673_{\color{red}{\uparrow{0.130}}}$ \\
				
				\cdashline{2-4}[0.8pt/2pt]\cdashline{5-7}[0.8pt/2pt]
				\specialrule{0em}{1pt}{1pt}
				&\cmark & \cmark & \cmark & $\textbf{0.621}_{\color{red}{\uparrow{0.085}}}$ & $\textbf{0.507}_{\color{red}{\uparrow{0.140}}}$ & $\textbf{0.895}_{\color{red}{\uparrow{0.352}}}$ \\				
				\bottomrule
		\end{tabular}}
		\vspace{-2mm}
		\caption{Ablation studies of the CPSTN and MRRN on TNO and RoadScene datasets.}
		\label{tab:our_ablation}
	\end{center}
	\vspace{-4mm}
\end{table}

\noindent
\textbf{MRRN.}
Suppose that in the absence of CPSTN, we investigate the effect of MRRN on misaligned IVIF tasks. Note that MRRN performs cross-modality image registration taking the paired infrared and visible images as inputs under the same settings.
As reported in Table~\ref{tab:our_ablation}, we compare the results of the first two experiments on the TNO and RoadScene datasets, which show that using MRRN to implement cross-modality registration also improves the performance of IVIF to a certain extent, compared with straightly fusing the misaligned infrared and visible images.
\vspace{1.0mm}

\begin{figure}[t]
	\footnotesize
	\begin{center}
		\begin{tabular}{ccccc}
			\multicolumn{3}{c}{\multirow{5}*[27.0pt]{\includegraphics[width=0.38\linewidth, height = 0.33\linewidth]{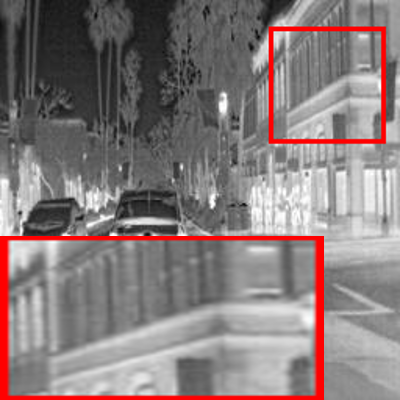}}}&\hspace{-4mm}
			\includegraphics[width=0.280\linewidth, height = 0.138\linewidth]{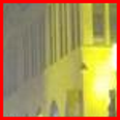} &\hspace{-4mm}
			\includegraphics[width=0.280\linewidth, height = 0.138\linewidth]{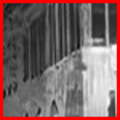} \\
			\multicolumn{3}{c}{~} &\hspace{-4mm}  \small(b) VIS &\hspace{-4mm}  (c) w/o \tiny$\mathcal{L}_{cross}$,$\mathcal{L}_{pst}$ \\
			\multicolumn{3}{c}{~} & \hspace{-4mm}
			\includegraphics[width=0.280\linewidth, height = 0.138\linewidth]{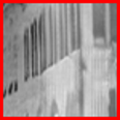} & \hspace{-4mm}
			\includegraphics[width=0.280\linewidth, height = 0.138\linewidth]{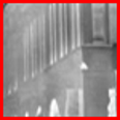}  \\
			
			\multicolumn{3}{c}{\hspace{-4mm} \small(a) Reference} &  \hspace{-4mm} \small(d) w/o $\mathcal{L}_{pst}$ &\hspace{-4mm}  \small(e) Ours \\
		\end{tabular}
	\end{center}
	\vspace{-4mm}
	\caption{Ablation analysis of loss functions in CPSTN on RoadScene dataset. }%
	\label{fig: ab-loss}
	\vspace{-3.6mm}
\end{figure}

\begin{figure}[t]
	\footnotesize
	\begin{center}
		\vspace{-2mm}
		\begin{tabular}{ccccc}			
			\multicolumn{3}{c}{\multirow{5}*[25.0pt]{\includegraphics[width=0.38\linewidth, height = 0.32\linewidth]{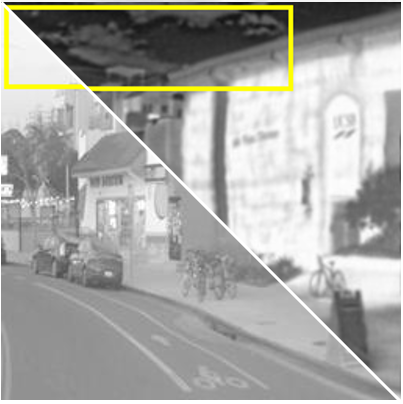}}}&\hspace{-4.5mm}
			\includegraphics[width=0.280\linewidth, height = 0.138\linewidth]{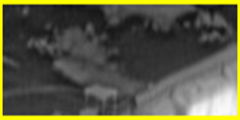} &\hspace{-4.5mm}
			\includegraphics[width=0.280\linewidth, height = 0.138\linewidth]{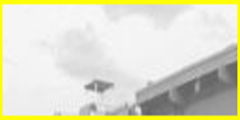} \\
			\multicolumn{3}{c}{~} &\hspace{-4mm}  \small(b) IR &\hspace{-4mm}  \small(c) VIS \\
			\multicolumn{3}{c}{~} & \hspace{-4mm}
			\includegraphics[width=0.280\linewidth, height = 0.138\linewidth]{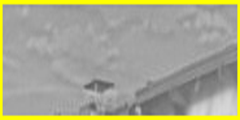} & \hspace{-4.5mm}
			\includegraphics[width=0.280\linewidth, height = 0.138\linewidth]{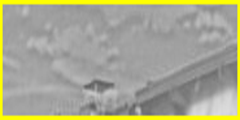}  \\
			
			\multicolumn{3}{c}{\hspace{-4mm} \small(a) VIS/IR} &  \hspace{-4mm} \small(d) w/o IFM &\hspace{-4mm}  \small(e) w/ IFM \\
		\end{tabular}
	\end{center}
	\vspace{-4mm}
	\caption{Ablation analysis of the IFM on RoadScene dataset. }%
	\label{fig: ab-ifm}
	\vspace{-4mm}
\end{figure}

\noindent
\textbf{Loss Functions in CPSTN.}
We analyze the effect of $\mathcal{L}_{pst}$ and $\mathcal{L}_{cross}$ used in the CPSTN. As shown in Figure~\ref{fig: ab-loss}, without using the constraints from $\mathcal{L}_{pst}$ and $\mathcal{L}_{cross}$, the generated pseudo infrared image suffers serious structural degradation (Figure~\ref{fig: ab-loss}(c)) compared with the reference image (Figure~\ref{fig: ab-loss}(a)). Using $\mathcal{L}_{cross}$, the model preserves general structural information, while subtle structures are not maintained well enough and "checkerboard artifacts" are introduced obviously (Figure~\ref{fig: ab-loss}(d)).
In contrast, the pseudo infrared image generated by our model (Figure~\ref{fig: ab-loss}(e)) has a sharper geometry structure, which caters to the common sense that infrared image "emphasizes structure over texture". 
\vspace{1.0mm}

\noindent
\textbf{Interaction Fusion Module.}
We replace the IFM in the DIFN by concatenation operation to examine its effectiveness for the IVIF. As observed in Figure~\ref{fig: ab-ifm}, the model without using the IFM tends to generate smooth textures with low contrast, while the result using the IFM has richer and sharper textures with positively high contrast. It is indicated that adaptively selecting features to be fused is more effective for improving fused image quality.

\subsection{Additional Analysis}

\begin{figure}[tbh!]
	\footnotesize
	\begin{center}
		\begin{tabular}{c}
			\hspace{-2.5mm}
			\includegraphics[width = 0.85\linewidth,height=0.14\textheight]{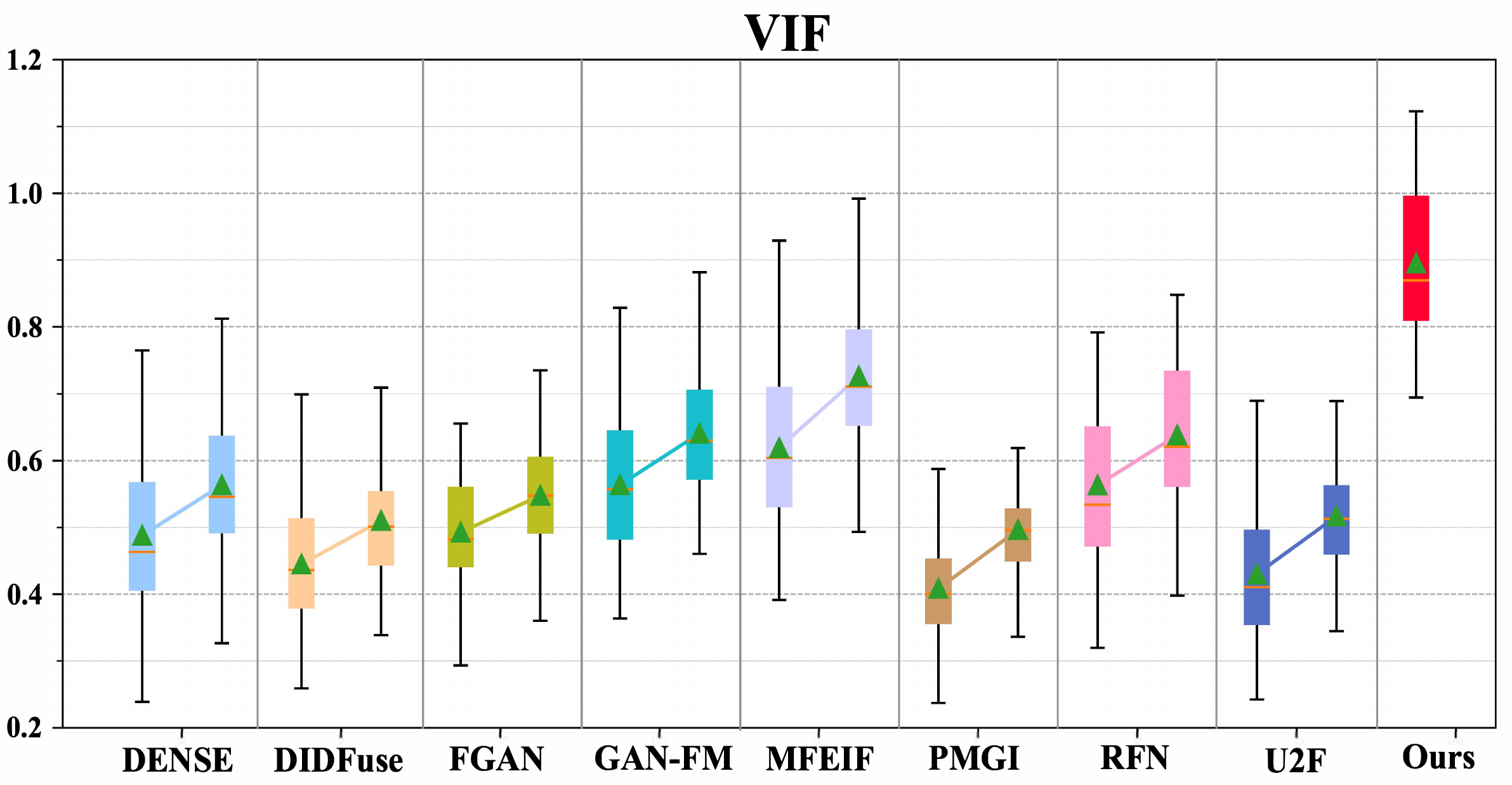} \\
			\hspace{-2.5mm}
			\includegraphics[width = 0.85\linewidth,height=0.14\textheight]{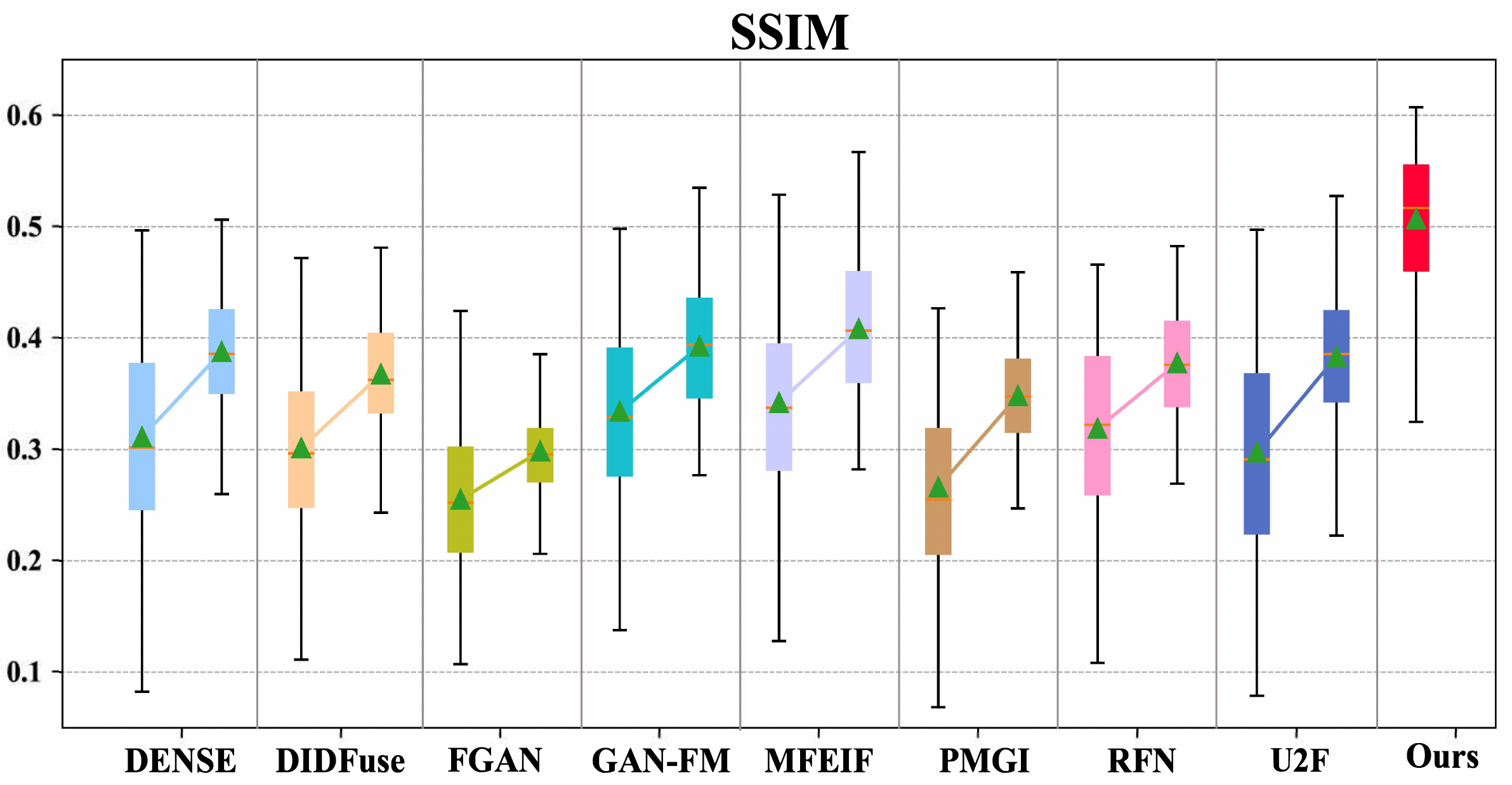}\\
		\end{tabular}
	\end{center}
	\vspace{-4mm}
	\caption{Generalization analysis of the proposed CPSTN. }
	\label{fig:road_with_T}
	\vspace{-4mm}	
\end{figure}

To verify the generalization of the proposed CPSTN, we plug it into two typical image alignment algorithms (i.e., FlowNet+STN and VoxelMorph) implement mono-modality image registration between distorted and pseudo infrared images, and further to implement fusion of the registered infrared image and visible image on eight existing IVIF methods. 
As shown in Figure~\ref{fig:road_with_T}, the first box of each method is the result of the model without using the CPSTN, and the second box is that of the model using the CPSTN.
It can be observed that: i) These alignment models equipped with the CPSTN can promote effectively the fusion performance of existing IVIF methods for misaligned images, which reveals the favorable generalization ability of the CPSTN for misaligned IVIF tasks.
ii) Although conditioned by i), the proposed method still outperforms these improved IVIF methods on misaligned cross-modality images, which explains that the superiority of our overall framework relies on the close cooperation of each component.


\section{Conclusion}
In this work, we proposed a highly-robust unsupervised misaligned infrared and visible image fusion framework for mitigating ghosts of the fused images. 
%
We exploited a generation-registration paradigm to simplify the cross-modality image alignment into a mono-modality registration.
%
%
A feature interaction fusion module was developed to adaptively select meaningful features from infrared and visible images to fuse, avoiding feature smoothing and emphasizing faithful textures. 
Extensive experimental results demonstrated the superior capability of our method on misaligned cross-modality image fusion.
Significantly, our generation-registration paradigm can be well extended to existing IVIF methods to improve their fusion performance on misaligned cross-modality images.

\section*{Acknowledgments}
This work is partially supported by the National Natural Science Foundation of China (Nos. 61922019, 61733002 and 62027826), and the Fundamental Research Funds for the Central Universities.

\bibliographystyle{named}
\bibliography{ijcai22}

\begin{thebibliography}{}

\bibitem[\protect\citeauthoryear{Arar \bgroup \em et al.\egroup }{2020}]{NeMAR}
Moab Arar, Yiftach Ginger, Dov Danon, Amit~H. Bermano, and Daniel Cohen{-}Or.
\newblock Unsupervised multi-modal image registration via geometry preserving
  image-to-image translation.
\newblock In {\em CVPR}, pages 13407--13416, 2020.

\bibitem[\protect\citeauthoryear{Balakrishnan \bgroup \em et al.\egroup
  }{2018}]{voxelmorph}
Guha Balakrishnan, Amy Zhao, Mert~R. Sabuncu, John~V. Guttag, and Adrian~V.
  Dalca.
\newblock An unsupervised learning model for deformable medical image
  registration.
\newblock In {\em CVPR}, pages 9252--9260, 2018.

\bibitem[\protect\citeauthoryear{Cao \bgroup \em et al.\egroup
  }{2018}]{NCC_metric}
Xiaohuan Cao, Jianhuan Yang, Li~Wang, Zhong Xue, Qian Wang, and Dinggang Shen.
\newblock Deep learning based inter-modality image registration supervised by
  intra-modality similarity.
\newblock In {\em MLMI}, pages 55--63, 2018.

\bibitem[\protect\citeauthoryear{Ghosh \bgroup \em et al.\egroup }{2019}]{SvSW}
Sanjay Ghosh, Ruturaj~G. Gavaskar, and Kunal~N. Chaudhury.
\newblock Saliency guided image detail enhancement.
\newblock In {\em NCC}, pages 1--6, 2019.

\bibitem[\protect\citeauthoryear{Han \bgroup \em et al.\egroup }{2013}]{vif}
Yu~Han, Yunze Cai, Yin Cao, and Xiaoming Xu.
\newblock A new image fusion performance metric based on visual information
  fidelity.
\newblock {\em Information Fusion}, 14(2):127--135, 2013.

\bibitem[\protect\citeauthoryear{Ilg \bgroup \em et al.\egroup
  }{2017}]{flownet}
Eddy Ilg, Nikolaus Mayer, Tonmoy Saikia, Margret Keuper, Alexey Dosovitskiy,
  and Thomas Brox.
\newblock Flownet 2.0: Evolution of optical flow estimation with deep networks.
\newblock In {\em CVPR}, pages 1647--1655, 2017.

\bibitem[\protect\citeauthoryear{Isola \bgroup \em et al.\egroup
  }{2017}]{pix2pix}
Phillip Isola, Jun{-}Yan Zhu, Tinghui Zhou, and Alexei~A. Efros.
\newblock Image-to-image translation with conditional adversarial networks.
\newblock In {\em CVPR}, pages 5967--5976, 2017.

\bibitem[\protect\citeauthoryear{Jaderberg \bgroup \em et al.\egroup
  }{2015}]{STN}
Max Jaderberg, Karen Simonyan, Andrew Zisserman, and Koray Kavukcuoglu.
\newblock Spatial transformer networks.
\newblock In {\em NeurIPS}, pages 2017--2025, 2015.

\bibitem[\protect\citeauthoryear{Lai \bgroup \em et al.\egroup
  }{2019}]{CharLoss}
Weisheng Lai, Jiabin Huang, Narendra Ahuja, and Ming{-}Hsuan Yang.
\newblock Fast and accurate image super-resolution with deep laplacian pyramid
  networks.
\newblock {\em {IEEE} Transactions on Pattern Analysis and Machine
  Intelligence}, 41(11):2599--2613, 2019.

\bibitem[\protect\citeauthoryear{Li and Wu}{2019}]{DenseFuse}
Hui Li and Xiao{-}Jun Wu.
\newblock Densefuse: {A} fusion approach to infrared and visible images.
\newblock {\em {IEEE} Transactions on Image Processing}, 28(5):2614--2623,
  2019.

\bibitem[\protect\citeauthoryear{Li \bgroup \em et al.\egroup }{2021}]{RFN}
Hui Li, Xiao{-}Jun Wu, and Josef Kittler.
\newblock Rfn-nest: An end-to-end residual fusion network for infrared and
  visible images.
\newblock {\em Information Fusion}, 73:72--86, 2021.

\bibitem[\protect\citeauthoryear{Liu \bgroup \em et al.\egroup
  }{2020a}]{LiuLZFL20}
Risheng Liu, Zi~Li, Yuxi Zhang, Xin Fan, and Zhongxuan Luo.
\newblock Bi-level probabilistic feature learning for deformable image
  registration.
\newblock In {\em IJCAI}, pages 723--730, 2020.

\bibitem[\protect\citeauthoryear{Liu \bgroup \em et al.\egroup
  }{2020b}]{liu2020bilevel}
Risheng Liu, Jinyuan Liu, Zhiying Jiang, Xin Fan, and Zhongxuan Luo.
\newblock A bilevel integrated model with data-driven layer ensemble for
  multi-modality image fusion.
\newblock {\em IEEE Transactions on Image Processing}, 30:1261--1274, 2020.

\bibitem[\protect\citeauthoryear{Liu \bgroup \em et al.\egroup }{2021a}]{MFEIF}
Jinyuan Liu, Xin Fan, Ji~Jiang, Risheng Liu, and Zhongxuan Luo.
\newblock Learning a deep multi-scale feature ensemble and an edge-attention
  guidance for image fusion.
\newblock {\em {IEEE} Transactions on Circuits and Systems for Video
  Technology}, 2021.

\bibitem[\protect\citeauthoryear{Liu \bgroup \em et al.\egroup
  }{2021b}]{SMoA_21}
Jinyuan Liu, Yuhui Wu, Zhanbo Huang, Risheng Liu, and Xin Fan.
\newblock Smoa: Searching a modality-oriented architecture for infrared and
  visible image fusion.
\newblock {\em {IEEE} Signal Processing Letters}, 28:1818--1822, 2021.

\bibitem[\protect\citeauthoryear{Liu \bgroup \em et al.\egroup
  }{2021c}]{liu2021investigating}
Risheng Liu, Jiaxin Gao, Jin Zhang, Deyu Meng, and Zhouchen Lin.
\newblock Investigating bi-level optimization for learning and vision from a
  unified perspective: A survey and beyond.
\newblock {\em IEEE Transactions on Pattern Analysis and Machine Intelligence},
  2021.

\bibitem[\protect\citeauthoryear{Liu \bgroup \em et al.\egroup
  }{2021d}]{Liu_pami_21}
Risheng Liu, Zi~Li, Xin Fan, Chenying Zhao, Hao Huang, and Zhongxuan Luo.
\newblock Learning deformable image registration from optimization:
  Perspective, modules, bilevel training and beyond.
\newblock {\em {IEEE} Transactions on Pattern Analysis and Machine
  Intelligence}, 2021.

\bibitem[\protect\citeauthoryear{Liu \bgroup \em et al.\egroup
  }{2021e}]{liu2021searching}
Risheng Liu, Zhu Liu, Jinyuan Liu, and Xin Fan.
\newblock Searching a hierarchically aggregated fusion architecture for fast
  multi-modality image fusion.
\newblock In {\em ACM MM}, pages 1600--1608, 2021.

\bibitem[\protect\citeauthoryear{Liu \bgroup \em et al.\egroup
  }{2022a}]{TarDAL_22}
Jinyuan Liu, Xin Fan, Zhangbo Huang, Guanyao Wu, Risheng Liu, Wei Zhong, and
  Zhongxuan Luo.
\newblock Target-aware dual adversarial learning and a multi-scenario
  multi-modality benchmark to fuse infrared and visible for object detection.
\newblock In {\em CVPR}, 2022.

\bibitem[\protect\citeauthoryear{Liu \bgroup \em et al.\egroup
  }{2022b}]{liu2022attention}
Jinyuan Liu, Jingjie Shang, Risheng Liu, and Xin Fan.
\newblock Attention-guided global-local adversarial learning for
  detail-preserving multi-exposure image fusion.
\newblock {\em IEEE Transactions on Circuits and Systems for Video Technology},
  2022.

\bibitem[\protect\citeauthoryear{Liu \bgroup \em et al.\egroup
  }{2022c}]{liu2022general}
Risheng Liu, Pan Mu, Xiaoming Yuan, Shangzhi Zeng, and Jin Zhang.
\newblock A general descent aggregation framework for gradient-based bi-level
  optimization.
\newblock {\em IEEE Transactions on Pattern Analysis and Machine Intelligence},
  2022.

\bibitem[\protect\citeauthoryear{Ma \bgroup \em et al.\egroup }{2019}]{FGAN}
Jiayi Ma, Wei Yu, Pengwei Liang, Chang Li, and Junjun Jiang.
\newblock Fusiongan: {A} generative adversarial network for infrared and
  visible image fusion.
\newblock {\em Information Fusion}, 48:11--26, 2019.

\bibitem[\protect\citeauthoryear{Mahapatra \bgroup \em et al.\egroup
  }{2018}]{MI_metric}
Dwarikanath Mahapatra, Zongyuan Ge, Suman Sedai, and Rajib Chakravorty.
\newblock Joint registration and segmentation of xray images using generative
  adversarial networks.
\newblock In {\em MICCAI}, pages 73--80, 2018.

\bibitem[\protect\citeauthoryear{Sajjadi \bgroup \em et al.\egroup
  }{2017}]{GramMatrix}
Mehdi S.~M. Sajjadi, Bernhard Sch{\"{o}}lkopf, and Michael Hirsch.
\newblock Enhancenet: Single image super-resolution through automated texture
  synthesis.
\newblock In {\em ICCV}, pages 4501--4510, 2017.

\bibitem[\protect\citeauthoryear{Simonyan and Zisserman}{2015}]{vgg}
Karen Simonyan and Andrew Zisserman.
\newblock Very deep convolutional networks for large-scale image recognition.
\newblock In {\em ICLR}, 2015.

\bibitem[\protect\citeauthoryear{Tang \bgroup \em et al.\egroup
  }{2020}]{BlockMix_20}
Hao Tang, Zechao Li, Zhimao Peng, and Jinhui Tang.
\newblock Blockmix: Meta regularization and self-calibrated inference for
  metric-based meta-learning.
\newblock In {\em ACM MM}, pages 610--618, 2020.

\bibitem[\protect\citeauthoryear{Tang \bgroup \em et al.\egroup
  }{2022}]{TANG2022108792}
Hao Tang, Chengcheng Yuan, Zechao Li, and Jinhui Tang.
\newblock Learning attention-guided pyramidal features for few-shot
  fine-grained recognition.
\newblock {\em Pattern Recognition}, 2022.

\bibitem[\protect\citeauthoryear{Wang \bgroup \em et al.\egroup }{2004}]{ssim}
Zhou Wang, Alan~C. Bovik, Hamid~R. Sheikh, and Eero~P. Simoncelli.
\newblock Image quality assessment: from error visibility to structural
  similarity.
\newblock {\em {IEEE} Transactions on Image Processing}, 13(4):600--612, 2004.

\bibitem[\protect\citeauthoryear{Wang \bgroup \em et al.\egroup
  }{2019a}]{AlignGAN}
Guan'an Wang, Tianzhu Zhang, Jian Cheng, Si~Liu, Yang Yang, and Zengguang Hou.
\newblock Rgb-infrared cross-modality person re-identification via joint pixel
  and feature alignment.
\newblock In {\em ICCV}, pages 3622--3631, 2019.

\bibitem[\protect\citeauthoryear{Wang \bgroup \em et al.\egroup }{2019b}]{D2RL}
Zhixiang Wang, Zheng Wang, Yinqiang Zheng, Yung{-}Yu Chuang, and Shin'ichi
  Satoh.
\newblock Learning to reduce dual-level discrepancy for infrared-visible person
  re-identification.
\newblock In {\em CVPR}, pages 618--626, 2019.

\bibitem[\protect\citeauthoryear{Xu \bgroup \em et al.\egroup
  }{2022}]{U2Fusion}
Han Xu, Jiayi Ma, Junjun Jiang, Xiaojie Guo, and Haibin Ling.
\newblock U2fusion: {A} unified unsupervised image fusion network.
\newblock {\em {IEEE} Transactions of Pattern Analysis and Machine
  Intelligence}, 44(1):502--518, 2022.

\bibitem[\protect\citeauthoryear{Yang \bgroup \em et al.\egroup
  }{2020}]{medireg}
Qianye Yang, Nannan Li, Zixu Zhao, Xingyu Fan, Eric~I{-}Chao Chang, and Yan Xu.
\newblock Mri cross-modality image-to-image translation.
\newblock {\em Scientific reports}, 10(1):1--18, 2020.

\bibitem[\protect\citeauthoryear{Zhang \bgroup \em et al.\egroup
  }{2021a}]{GAN-FM}
Hao Zhang, Jiteng Yuan, Xin Tian, and Jiayi Ma.
\newblock {GAN-FM:} infrared and visible image fusion using {GAN} with
  full-scale skip connection and dual markovian discriminators.
\newblock {\em {IEEE} Transactions on Computational Imaging}, 7:1134--1147,
  2021.

\bibitem[\protect\citeauthoryear{Zhang \bgroup \em et al.\egroup }{2021b}]{RDN}
Yulun Zhang, Yapeng Tian, Yu~Kong, Bineng Zhong, and Yun Fu.
\newblock Residual dense network for image restoration.
\newblock {\em {IEEE} Transactions on Pattern Analysis and Machine
  Intelligence}, 43(7):2480--2495, 2021.

\bibitem[\protect\citeauthoryear{Zhao \bgroup \em et al.\egroup
  }{2020}]{DIDFuse_2020}
Zixiang Zhao, Shuang Xu, Chunxia Zhang, Junmin Liu, Jiangshe Zhang, and Pengfei
  Li.
\newblock Didfuse: Deep image decomposition for infrared and visible image
  fusion.
\newblock In {\em IJCAI}, pages 970--976, 2020.

\bibitem[\protect\citeauthoryear{Zhu \bgroup \em et al.\egroup
  }{2017}]{cyclegan}
Jun{-}Yan Zhu, Taesung Park, Phillip Isola, and Alexei~A. Efros.
\newblock Unpaired image-to-image translation using cycle-consistent
  adversarial networks.
\newblock In {\em ICCV}, pages 2242--2251, 2017.

\end{thebibliography}

\end{document}